\documentclass{article}
\PassOptionsToPackage{numbers, compress}{natbib}
\usepackage[preprint]{neurips_2026}
\usepackage{booktabs}
\usepackage{multirow}
\usepackage[utf8]{inputenc}
\usepackage[T1]{fontenc}
\usepackage{hyperref}
\usepackage{url}
\usepackage{amsfonts}
\usepackage{nicefrac}
\usepackage{microtype}
\usepackage{xcolor}
\usepackage{graphicx}
\usepackage{amsmath}
\usepackage{amssymb}
\usepackage{mathtools}
\usepackage{subcaption}
\usepackage{enumitem}
\usepackage{float}
\usepackage{listings}
\usepackage{tabularx}
\usepackage{array}
\usepackage{ragged2e}

\usepackage{tcolorbox}
\usepackage{needspace}
\usepackage{pgfplots}
\tcbuselibrary{breakable}

\pgfplotsset{compat=1.18}
\tcbuselibrary{breakable,skins,listings}

\definecolor{promptbg}{RGB}{248,249,251}
\definecolor{promptframe}{RGB}{210,214,220}
\definecolor{prompttitle}{RGB}{60,70,90}

\definecolor{citeblue}{RGB}{0,80,180}

\hypersetup{
    colorlinks=true,
    citecolor=citeblue,
    linkcolor=citeblue,
    urlcolor=citeblue
}
\lstdefinestyle{promptstyle}{
    basicstyle=\ttfamily\footnotesize,
    breaklines=true,
    breakatwhitespace=false,
    columns=fullflexible,
    keepspaces=true,
    showstringspaces=false,
    frame=none,
    xleftmargin=0.4em,
    xrightmargin=0.4em,
    aboveskip=0pt,
    belowskip=0pt,
    numbers=left,
    numberstyle=\scriptsize\color{gray},
    numbersep=8pt,
    tabsize=2
}

\newtcblisting{promptbox}[2][]{%
    enhanced,
    breakable,
    colback=promptbg,
    colframe=promptframe,
    boxrule=0.5pt,
    arc=2pt,
    left=4pt,
    right=4pt,
    top=4pt,
    bottom=4pt,
    title={\textbf{#2}},
    coltitle=prompttitle,
    fonttitle=\small,
    listing only,
    before=\par\Needspace{20\baselineskip},
    listing options={style=promptstyle,#1}
}

\newcommand{\method}{\textsc{H-Mem}}

\title{\method{}: A Novel Memory Mechanism for Evolving and Retrieving Agent Memory via a Hybrid Structure}

\author{
  \textbf{Jiawei Yu}$^{1}$,
  \textbf{Yixiang Fang}$^{1}$\thanks{Corresponding author.},
  \textbf{Xilin Liu}$^{2}$,
  \textbf{Yuchi Ma}$^{2}$ \\
  $^{1}$The Chinese University of Hong Kong, Shenzhen \\
  $^{2}$Huawei Cloud Computing Technologies CO., LTD. \\
  \texttt{jiaweiyu1@link.cuhk.edu.cn, fangyixiang@cuhk.edu.cn} \\
  \texttt{\{liuxilin3, mayuchi1\}@huawei.com}
}

\begin{document}

\maketitle

\begin{abstract}
Memory data are ubiquitous in Large Language Model (LLM)-based agents (e.g., OpenClaw and Manus).
A few recent works have attempted to exploit agents' memory for improving their performance on the question-answering (QA) task, but they lack a principled mechanism for effectively modeling how memory data evolves over time and retrieving memory data effectively, leading to poor performance in memory utilization.
To fill this gap, we present \method{}, a novel memory mechanism via a hybrid structure that can not only effectively model the evolution of agent memory over a long period of time, but also provide an efficient memory retrieval approach.
Particularly, \method{} builds a temporal and semantic tree structure that allows the short-term memory data to evolve progressively into long-term memory data, where the latter provides summarized information about the former, while simultaneously constructing a knowledge graph to capture the relationships between entities in memory.
Moreover, it offers an effective memory retrieval approach by exploiting the hybrid structure of the tree and graph structures.
Extensive experiments on three agent memory benchmarks show that \method{} achieves state-of-the-art performance on the QA task.
\end{abstract}
\section{Introduction}
\label{sec:intro}

LLM-based agents such as OpenClaw~\citep{openclaw2026github} and Manus~\citep{manus2025official} have received tremendous attention owing to their powerful abilities in solving complex real-world tasks such as QA.
During the interactions between users and agents, a large amount of memory data has been generated and accumulated.
Generally, the agent memory data refers to the information accumulated by an agent during interactions, such as conversation history and task execution records.
By exploiting the memory data, the agent can not only have a clear understanding of the users' preferences and behaviors but also improve performance on understanding context, maintaining conversational coherence, and executing complex tasks.
As a result, a crucial component of modern agents is the memory mechanism, which stores and manipulates the agent memory data.

To enable an LLM-based agent to exploit memory data, a naive memory mechanism is to store the memory as plain text, retrieve all memory data associated with a specific user, and then use the retrieved data to accomplish a task for that user.
However, due to the finite context window of LLMs, this mechanism cannot effectively or efficiently process large amounts of memory data, especially when the agent has interacted with users over a long period of time.
To alleviate this bottleneck, existing systems often apply Retrieval-Augmented Generation (RAG) techniques to memory data~\citep{lewis2020rag,wu2026memoryllmeramodular}; that is, the agent only retrieves relevant memory information from an external memory database when solving a task.

\par









\begin{table*}[h]
\centering
\small
\setlength{\tabcolsep}{2.8pt}
\renewcommand{\arraystretch}{1.12}
\caption{Taxonomy of representative agent memory methods.}
\label{tab:memory_taxonomy}
\resizebox{\textwidth}{!}{
\begin{tabular}{p{2.3cm} p{3.0cm} p{6cm} p{1.6cm} p{1.8cm}}
\toprule
\textbf{Method} 
& \textbf{Main Index Structures} 
& \textbf{Memory Retrieval Method}
& \textbf{Memory Evolution}
& \textbf{Multi-hop Reasoning} \\
\midrule

Mem0~\citep{chhikara2025mem0}
& Vector
& Vector-based
& No
& No \\

\midrule

MemOS~\citep{li2025memos}
& Tree
& Lexical-based and vector-based
& No
& No \\

MemTree~\citep{rezazadeh2024memtree}
& Tree
& Vector-based and structure-based
& No
& No \\

MemoryOS~\citep{kang2025memoryos}
& Hierarchical
& Lexical-based and vector-based
& No
& No \\

EverMemOS~\citep{hu2026evermemos}
& Hierarchical
& Lexical-based, vector-based, and agentic
& Yes
& No \\

\midrule

Zep~\citep{rasmussen2025zep}
& Graph
& Lexical-based, vector-based, and structure-based
& No
& Yes \\

\midrule

\method{}
& Hybrid of tree and graph
& Vector-based, structure-based, and agentic
& Yes
& Yes \\

\bottomrule
\end{tabular}
}
\end{table*}

Under this memory-based RAG paradigm, existing methods differ not only in how memory data are indexed before retrieval, but also in their retrieval mechanisms which determine how relevant evidence is accessed from the index.
According to the memory index structures, existing memory mechanisms can be roughly classified into three categories as reported in Table \ref{tab:memory_taxonomy}.
In Table~\ref{tab:memory_taxonomy}, \emph{Memory Evolution} denotes temporal-window-based consolidation from short-term memories to long-term summaries, while \emph{Multi-hop Reasoning} denotes entity- or relation-level traversal across memory fragments.
A first class of methods adopts the \emph{vector index}, a single-level organization of memory, in which memory fragments are stored as independent entries.
To enable efficient memory retrieval, these fragments are often encoded as embeddings, and some methods further store these embeddings in a vector database~\citep{zhong2024memorybank,chhikara2025mem0}.
%
%
A second class of methods mainly explores the \emph{tree index}, where the semantic topics of the memory data are hierarchically organized across multiple levels, with lower levels preserving fine-grained semantic topics and higher levels providing the abstract or persistent representations of fine-grained semantic topics
~\citep{rezazadeh2024memtree,kang2025memoryos,li2025memos,hu2026evermemos}.
To query the memory fragments about a specific topic, they just need to traverse explicitly along the tree structure in a bottom-up or top-down manner.
%
A third class of methods mainly uses the \emph{graph index}, where entities and relationships are represented as nodes and edges, respectively~\citep{rasmussen2025zep}.
By following the link relationships between entities, they can naturally support fast relational retrieval and multi-hop reasoning.

Despite this progress, existing memory mechanisms suffer from two major limitations:
First, their index designs are still limited in modeling memory evolution, where short-term memory can be progressively consolidated into long-term memory, as suggested by studies of human memory consolidation~\citep{mcgaugh2000memory,squire1995retrograde}.
This is primarily because they fail to explicitly take the temporal dimension into account, which renders them incapable of differentiating between short- and long-term semantic topics within the memory data.
Second, they cannot accurately retrieve the relevant evidence from the memory index when performing QA tasks. 
Specifically, the vector index-based methods are efficient for similarity search, but they treat memory data as independent entries and therefore cannot explicitly capture either temporal abstraction or entity-level relational dependencies.
The tree index-based methods cannot accurately capture the multi-hop relationships between entities;
and the graph index-based methods cannot identify the consolidated memory data due to the lack of a memory evolution mechanism.
Overall, these methods mainly rely on a single index (i.e., vector, tree, or graph), so they cannot accurately retrieve the relevant evidence from memory data.
Thus, existing works lack a principled mechanism that can jointly model long-term memory evolution and support accurate retrieval.

To address the aforementioned limitations, we propose \method{}, a novel memory mechanism via a hybrid structure of tree and graph.
%
%
The key distinction of \method{} is not merely using a tree index together with a graph index, but coupling temporal-semantic memory evolution with entity-centered multi-hop reasoning.
The tree structure of \method{} organizes memory data both temporally and semantically, where each tree node retains memory information regarding a specific semantic topic within a pre-defined time window. 
Specifically, each leaf node stores an event of the agent's original memory fragment, containing a semantic topic (e.g., a message in a conversation) generated at a specific timestamp, while the upper-level nodes store the memory summaries of fine-grained semantic topics in their lower levels, covering their respective time windows.
To enable memory evolution, \method{} performs a temporal-and-semantic consolidation; that is, given two tree nodes whose time windows are very close in the same level, if the semantic similarity between their memory data exceeds a predefined threshold, they could share the same parent node, whose memory summary preserves the consolidated information of these two nodes. 
Clearly, this temporal and semantic tree structure allows short-term memory to evolve progressively into long-term memory.
Furthermore, the graph structure of \method{} maintains a knowledge graph of entities and their relationships extracted from the memory data, effectively recording the entity-centered information beyond temporal order and capturing multi-hop relationships between entities across different memory fragments.
Overall, the tree and graph structures complement each other, and this hybrid structure overcomes the issue of relying on a single index prevalent in existing works.


Based on this hybrid structure, \method{} includes an effective retrieval method.
Given a query $Q$, it first decomposes $Q$ into some sub-queries and generates a retrieval workflow for each sub-query.
Then, for each sub-query, it locates some original memory fragments and multi-hop relevant entities in the graphs.
Afterwards, it searches relevant evidence from the tree in a bottom-up manner, which is used for completing the RAG process.
We have evaluated \method{} against representative SOTA baselines on three public long-term memory benchmarks covering diverse QA scenarios.
The results show that \method{} achieves superior F1 scores and accuracy while maintaining competitive index and retrieval efficiency.
Further analyses validate the contribution of the temporal tree, the knowledge graph, and the agent-assisted retrieval strategy.

Our principal contributions are summarized as follows:
\begin{itemize}[nosep]
    \item We propose \method{}, a novel memory mechanism that can effectively model the evolution of agent memory over a long time by using a hybrid structure of tree and graph.
    
    \item Based on the hybrid structure above, we develop an effective method for retrieving the relevant memory evidence to support the QA tasks.
    
    \item Experiments on three public long-term agent memory benchmarks show that \method{} achieves SOTA performance in solving QA tasks while maintaining competitive efficiency.
\end{itemize}

\section{Related Work}
\label{sec:related}

\subsection{Retrieval-Augmented Generation (RAG)}
\label{sec:rag}

Recently, many works have explored how LLMs can access external information beyond its parametric knowledge and immediate prompt context.
Simply extending the context window is insufficient, as the key challenge is how to select, organize, and reuse the external information effectively.
Within this landscape, RAG has become a widely used technique for incorporating external knowledge at inference time~\citep{lewis2020rag,zhou2025depth}.
Given a question $Q$, it retrieves the relevant information from an external database, incorporates it with $Q$ as the prompt, and then feeds it into the LLM for generation.
Various types of RAG techniques have been studied: the naive RAG retrieves relevant passages from external corpora, graph-based RAG leverages a graph-structured index for multi-hop and relation-aware reasoning~\citep{edge2024graphrag, gutierrez2024hipporag}, and agentic RAG incorporates retrieval into an adaptive reasoning loop so that the model can decide when and how to retrieve during multi-step problem solving~\citep{asai2023selfrag, lee2024planrag}. 

\subsection{Agent Memory-based RAG and Agent Memory Mechanisms}
\label{sec:agentMemory}

Since the agent memory data can be considered as a kind of external information, it is natural to use it for RAG.
The memory-based RAG techniques \cite{wu2026memoryllmeramodular} often first extract useful information from memory data, such as user preferences and events, then organize them into some index structures, and finally retrieve relevant evidence and inject it into the prompt when answering a question.
However, different from traditional RAG techniques, which often use static documents to provide factual grounding, the memory-based RAG techniques operate over stateful, interaction-derived memory data that evolves over time, and aim to understand context, maintain conversational coherence, and execute complex tasks.
Therefore, they heavily rely on the memory mechanisms, which not only provide an effective organization of the memory data but also offer effective methods for evolving and retrieving the memory data.

According to the memory index structures, existing memory mechanisms can be roughly classified into three categories:
(1) The vector-based memory methods, such as MemoryBank~\citep{zhong2024memorybank} and Mem0~\citep{chhikara2025mem0}, store interaction-derived memory as independent embeddings and retrieve relevant memory from ongoing interactions.
(2) The tree-based memory methods, such as MemTree~\citep{rezazadeh2024memtree}, introduce dynamic tree-structured representations to organize memory at different abstraction levels.
MemOS~\citep{li2025memos} also supports tree-like textual memory modules within its MemCube abstraction.
Related hierarchical memory methods, such as MemoryOS~\citep{kang2025memoryos} and EverMemOS~\citep{hu2026evermemos}, also organize memories across multiple levels or structured units, emphasizing memory management and long-term reuse.
(3) The graph-based memory methods, such as Zep~\citep{rasmussen2025zep}, construct temporal knowledge graphs for agent memory, enabling relational access to evolving facts and entities.
Additionally, recent works have explored structured and adaptive memory mechanisms from related perspectives.
A-Mem~\citep{xu2025amem} studies agentic memory mechanisms, while multi-granularity memory methods~\citep{xu2025multigranularity} investigate memory association and selection across different abstraction levels.

As aforementioned, although the above works have achieved some promising progress, their index designs are still limited in modeling the evolution of memory data, which progressively consolidates short-term memory fragments into long-term memory fragments.
Besides, they cannot accurately retrieve the relevant evidence from the memory index when performing QA tasks. 
Therefore, it is desirable to study a novel memory mechanism that can not only effectively model the evolution of agent memory over a long period of time, but also provide an efficient memory retrieval approach.

\section{Our Proposed Memory Mechanism \method{}}
\label{sec:method}

To effectively support the memory-based RAG, we propose \method{}, a novel memory mechanism for evolving and retrieving agent memory.
\method{} consists of two stages: \textbf{Offline Indexing} and \textbf{Online Retrieval}, where the former stage builds a hybrid structure of tree and graph, and the latter stage includes an agentic memory retrieval approach by exploiting the hybrid structure.
%

\begin{figure*}[t]
    \centering
    \includegraphics[width=0.496\textwidth]{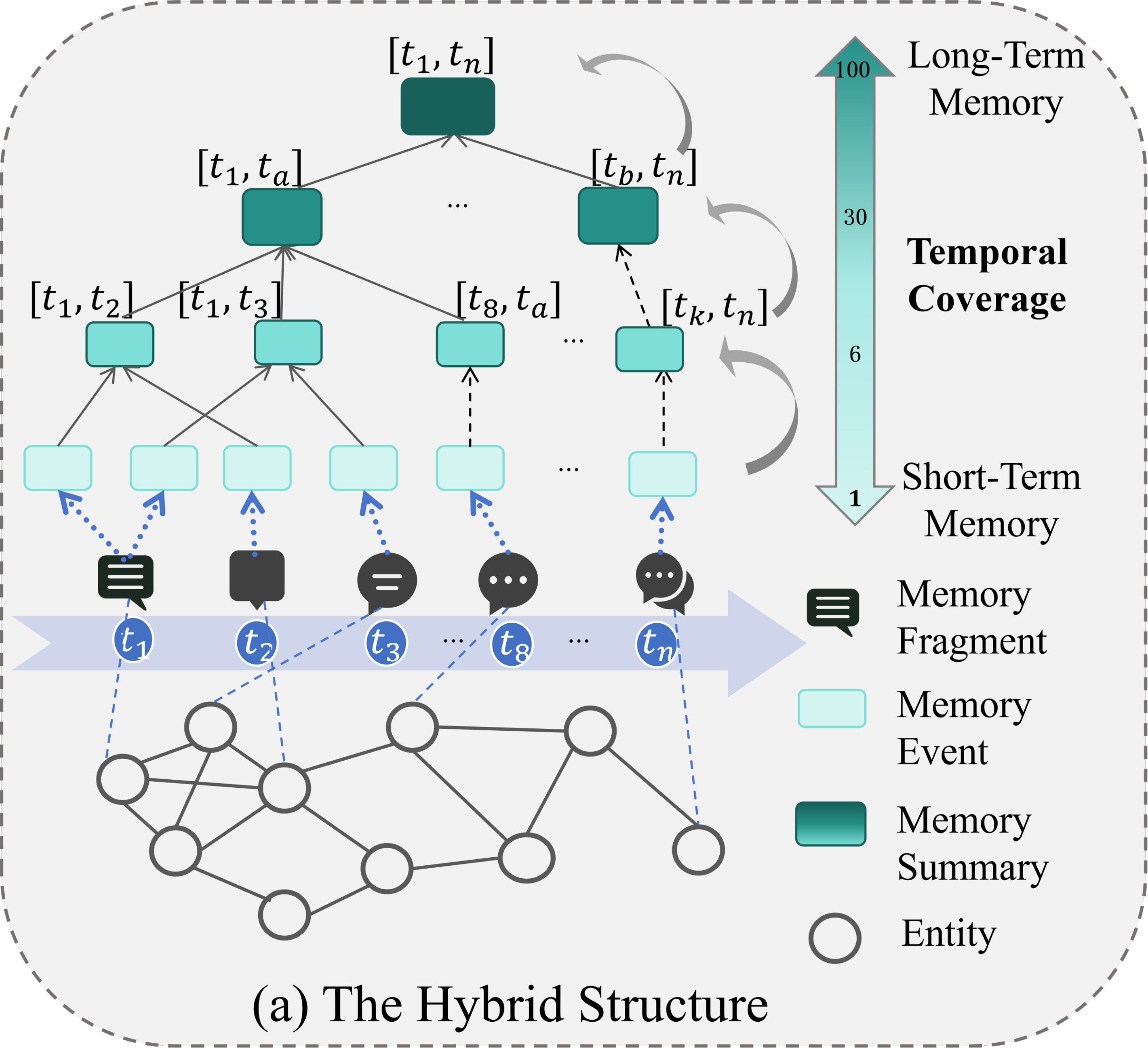}
    \hfill
    \includegraphics[width=0.496\textwidth]{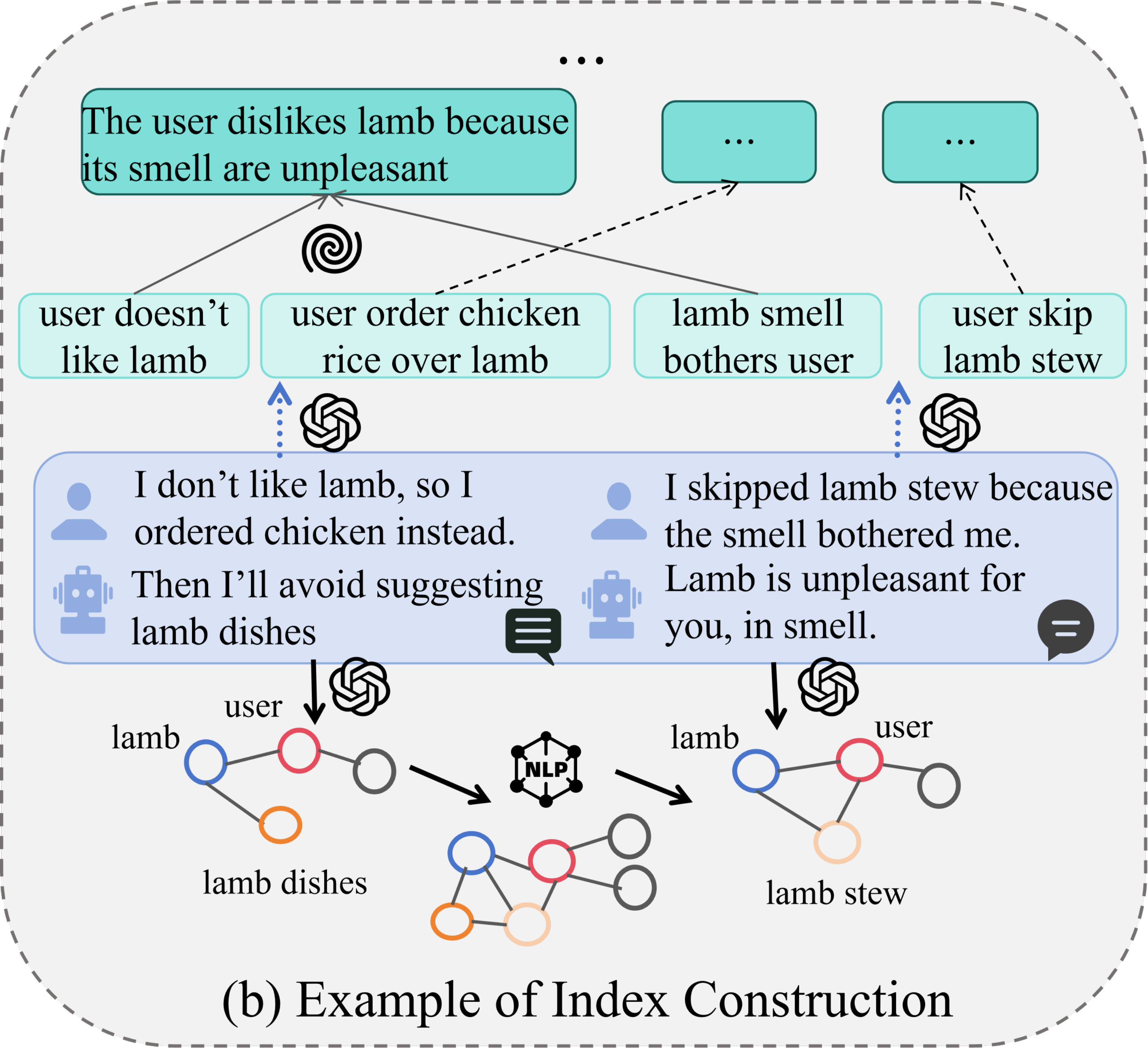}
    \caption{The offline indexing stage of \method{}.}
    \label{fig:index}
\end{figure*}

\subsection{Overview}
\label{sec:overview}

Let $\mathcal{F}=\{f_i\}$ be the set of original {\it memory fragments} of the memory data.
In the offline indexing stage, as depicted in Figure~\ref{fig:index}, \method{} builds a hybrid structure for $\mathcal F$, mainly consisting of two parts:
\begin{itemize}[nosep]
    

    \item {\bf Tree:} We build a temporal and semantic tree $\mathcal T$, where each node retains memory information regarding a specific semantic topic within a pre-defined time window. In the temporal view, for all the levels from the leaf to the root, their nodes have pre-defined time windows (e.g., one day, one week, one month, etc), and the time window of a parent node covers its child nodes' time windows. In the semantic view, each leaf node contains a memory event extracted from an original memory fragment, and each non-leaf node stores a {\it memory summary}, providing an abstract and persistent representation of the fine-grained semantic topics of memory events/summaries in its child nodes. 

    \item {\bf Graph:} We build a knowledge graph $\mathcal G=(\mathcal V, \mathcal E)$, where $\mathcal V$ contains the entities extracted from the original memory fragments and $\mathcal E$ denotes relationships between entities. Besides, each entity is linked to the original memory fragment containing it and may have a profile.
\end{itemize}

The key rationale behind our design is that the tree above can effectively support memory evolution in both temporal and semantic dimensions across different granularities, while the graph records the entity-centered information beyond the temporal dimension and captures multi-hop relationships between entities across different memory fragments.
Additionally, to enable efficient semantic search, we maintain the embedding vectors of the memory fragments/events/summaries.

In the online retrieval stage, \method{} first decomposes a query $Q$ into sub-queries.
Then for each of them, it searches the relevant memory information within some specific time windows by using the hybrid structure.
Finally, it combines the searched memory information for all the sub-queries as the relevant evidence identified from the memory data.

\subsection{Offline Indexing}
\label{sec:offline_indexing}

We now introduce the construction process of the hybrid structure.


{\bf Tree Construction.} 
We build a temporal and semantic tree $\mathcal T$ in an incremental manner.
Assume that $\mathcal T$ has $L$ levels, where the leaf nodes are at the first level, and each level $l$ has two hyperparameters $\alpha_l$ and $\beta_l$, where $\alpha_l$ is the similarity threshold between a node and its child node, and $\beta_l$ is the size of the time window.
As shown in Figure~\ref{fig:index}(a), given a list of original memory fragments which may include speakers, timestamps, and conversation data, we first extract a set of memory events from each memory fragment, where each memory event preserves a fine-grained semantic topic.
Next, we create a leaf node $x$ for each newly extracted memory event and update the tree in a bottom-up manner.
At each level $l$, the newly inserted node is assigned to the corresponding temporal window with size $\beta_l$ according to its timestamp.
Only the existing nodes within the same temporal window are considered as candidates for semantic consolidation.
Within this temporal window, we compute pairwise semantic similarities between the newly inserted node and existing lower-level nodes. 
If the newly inserted node is similar to an existing candidate cluster according to the threshold $\alpha_l$, we add it to this candidate cluster.
Then, we update a non-leaf node $y$ as the parent node of this candidate cluster in the upper level, and generate a memory summary by consolidating the memory events/summaries from its child nodes.
Otherwise, we create a new non-leaf node $z$ as the parent node of the newly inserted node in the upper level.
Afterwards, we repeat the above process by updating the parents of the newly generated node $y$ or $z$, until we reach the root node.
%
%
In this way, the tree supports memory evolution across different temporal windows; that is, the leaf nodes retain fine-grained short-term memory, while upper-level nodes provide abstract or persistent long-term memory.

{\bf Graph Construction.}
\method{} also incrementally builds a knowledge graph $\mathcal{G}$.
First, it extracts entities and relations from each original memory fragment $f_i$.
Second, the extracted entities are normalized and resolved into entity nodes through entity disambiguation based on text normalization, lemmatization, token overlap, and fuzzy string matching. 
If a resolved entity exactly matches an existing entity node and their types are compatible, it is merged into that node; otherwise, it is kept as a new entity node, and some associated edges may be inserted.
Besides, the merged or new entity node is linked to the original memory fragment containing it.
Third, the extracted relations are mapped to their resolved head and tail entity nodes and then inserted into the graph as an edge if the same edge does not already exist.
Therefore, the graph provides an entity-centered view of memory and captures multi-hop relations between entities across different memory fragments.
In addition, \method{} maintains profiles for salient entities selected based on the number of memory fragments--entity links associated with an entity, together with predefined important entity types such as persons, organizations, and locations.
For each salient entity, \method{} maintains a profile keeping both persistent and recent memory data.

Overall, \method{} constructs a hybrid structure with some vectors, a tree, and a graph that are built incrementally.
It can not only effectively model the evolution of agent memory over a long time, but also provide a structure supporting efficient memory retrieval introduced in the next subsection.

%



\subsection{Online Retrieval}
\label{sec:online_retrieval}

Given a query $Q$, \method{} identifies relevant memory evidence by searching over the hybrid structure, which consists of three steps, i.e., Retrieval Planning, Evidence Retrieval, and Generation Process.

{\bf 1) Retrieval Planning.}
\method{} first decomposes $Q$ into a list of sub-queries $\{Q_k\}_{k=1}^{K}$ with dependency relations in an agentic manner.
For each $Q_k$, it uses an LLM to infer a memory scope with a label in $\{\textsc{Short},\textsc{Long},\textsc{Mixed}\}$, indicating the temporal granularity for tree-based retrieval, where \textsc{Short} focuses on short-term memory evidence, \textsc{Long} focuses on long-term memory evidence, and \textsc{Mixed} considers both.
Besides, it infers explicit temporal hints if available, such as specific dates and relative time cues.
Finally, it generates a retrieval workflow for $Q_k$.

{\bf 2) Evidence Retrieval.}
Following the retrieval workflow, \method{} retrieves relevant evidence for each $Q_k$ by first exploring entities in the graph and then searching the tree in the hybrid memory structure.
Specifically, by extracting entities from $Q_k$, it locates seed entities in the graph through NLP-based lexical entity matching and vector-based semantic similarity search.
Starting from the seed entities, it then performs multi-hop expansion in the graph to identify more related entities.

Based on the identified entities, \method{} further explores the tree to identify relevant memory evidence in a bottom-up manner.
Specifically, it first maps the entities to their original memory fragments and then links them to memory events in the tree.
Afterwards, it searches the tree according to the inferred memory scope:
If the scope is \textsc{Short}, it only uses the original memory fragments and memory events in the leaf nodes of the tree;
If the scope is \textsc{Long}, it only uses the memory events and memory summaries in the tree;
and if the scope is \textsc{Mixed}, it uses the original memory fragments, memory events, and memory summaries.

For each memory evidence, denoted by $m$, \method{} considers three aspects between $m$ and $Q_k$: {\it semantic similarity}, {\it temporal relevance}, and {\it memory robustness}.
The semantic similarity as $S(m, Q_k)$ is derived by the cosine similarity between the embedding vectors of $m$ and $Q_k$.
For temporal relevance, let $I_m=[s_m,e_m]$ denote the time interval of memory evidence $m$, and $I_k=[s_k,e_k]$ denote the temporal interval inferred from the query $Q_k$.
The temporal relevance is computed by jointly considering temporal overlap and normalized center distance:
\begin{equation}
\small
T(m,Q_k)
=
\lambda \cdot
\frac{
\left| I_m \cap I_k \right|
}{
\left| I_m \cup I_k \right|+\epsilon
}
+
(1-\lambda)
\left(
1-
\frac{|c_m-c_k|}{\left| I_m \cup I_k \right|+\epsilon}
\right),
\end{equation}
where $c_m$ and $c_k$ are the centers of $I_m$ and $I_k$, $\lambda\in[0,1]$ controls the balance between temporal overlap and temporal distance, and $\epsilon$ is a small constant for numerical stability.

The memory robustness is used to reflect how likely a memory event or summary is retained during memory evolution.
Inspired by the Ebbinghaus forgetting curve~\citep{ebbinghaus1913memory}, we formulate the memory robustness of $m$ at the query time $t$ as
\begin{equation}
\small
R(m,t)=
\exp\!\left(
-\frac{t-r_m}{\tau(1+\eta\ln(1+n_m))}
\right),
\end{equation}
where $r_m$ is the most recent timestamp when $m$ is consolidated,
$\tau>0$ controls the forgetting time scale,
$\eta \ge 0$ controls the reinforcement effect of repeated mentions,
and $n_m$ is the number of times that $m$ is consolidated in the memory evolution.
A higher robustness value indicates that $m$ is more recent or has been repeatedly mentioned, suggesting that it is more likely to serve as reliable evidence.

Finally, all retrieved evidence is de-duplicated and ranked into an evidence chain by jointly considering semantic similarity, temporal relevance, and memory robustness as follows:
\begin{equation}
\small
\mathcal{F}(m,Q_k,t)
=
\theta_1 S(m,Q_k)
+
\theta_2 T(m,Q_k)
+
\theta_3 R(m,t),
\end{equation}
where $\theta_1$, $\theta_2$, and $\theta_3$ are non-negative weights, and $t$ is the query time of $Q$.
The evidence yielding higher $\mathcal{F}(m, Q_k,t)$ scores is prioritized to construct the final evidence chain for answering $Q$.

{\bf 3) Generation Process.}
For each $Q_k$, \method{} generates a sub-answer using its retrieved evidence. If $Q_k$ depends on other sub-queries, the answers to these prerequisite sub-queries are also used as additional context.
%
%
%
Finally, \method{} synthesizes all sub-query answers as $\Psi(\{\mathcal{A}_k\}_{k=1}^{K})$, where $\mathcal{A}_k$ denotes the answer of $Q_k$ and we invoke an LLM to complete the synthesis process $\Psi$.
\section{Experiments}
\label{sec:experiments}

We present the experimental setup in Section \ref{sec:experimental_setup} and discuss the results in Sections~\ref{sec:main_results} and \ref {sec:ablation}.

\subsection{Setup}
\label{sec:experimental_setup}

{\bf Datasets.}
We evaluate \method{} on three public long-term agent memory benchmarks: \textbf{LoCoMo}~\citep{maharana2024locomo}, \textbf{LongMemEvalS}~\citep{wu2024longmemeval}, and \textbf{REALTALK}~\citep{lee2025realtalk}. 
LoCoMo~\citep{maharana2024locomo} evaluates long-term conversational memory over ultra-long multi-session dialogues and contains 1,540 questions over 10 dialogues, covering single-hop, multi-hop, temporal, and open-domain questions. 
LongMemEvalS evaluates long-term interactive memory in assistant-style settings; in the S-setting, each conversation contains roughly 115K tokens, and the benchmark includes 500 questions spanning core capabilities. 
Both LoCoMo and LongMemEvalS are constructed in controlled LLM-simulated settings. 
In contrast, REALTALK is built from crowdsourced real-world human--human conversations, providing a more realistic testbed for persistent conversational memory. 
Together, these datasets cover both controlled and realistic settings, as well as diverse long-term memory demands.

{\bf Metrics.}
We consider two complementary evaluation metrics: \textbf{F1} and \textbf{LLM-Judge Accuracy}. 
F1 is computed from token-level precision and recall between the predicted answer and the reference answer, and measures partial lexical overlap.
Since long-term memory questions often admit semantically correct yet lexically diverse responses, we additionally report LLM-Judge Accuracy following prior long-term memory QA evaluation protocols~\citep{packer2024memgpt}, which marks a prediction as correct only if an LLM judge determines that it is semantically consistent with the gold answer.
This combination allows us to evaluate both lexical-level answer overlap and semantic correctness.

{\bf Baselines and Configurations.}
We consider six representative agent memory methods that reflect different design choices for agent memory: \text{MemoryOS}~\citep{kang2025memoryos}, \text{Mem0}~\citep{chhikara2025mem0},
\text{MemTree}~\citep{rezazadeh2024memtree}, \text{MemOS}~\citep{li2025memos}, \text{Zep}~\citep{rasmussen2025zep}, and \text{EverMemOS}~\citep{hu2026evermemos},
which are listed in Table \ref{tab:memory_taxonomy}.
%
%
To ensure a fair comparison, we evaluate \method{} and all baselines with the same embedding model, re-ranking model, and LLM-as-a-judge prompt.
This unified configuration reduces variance introduced by method-specific evaluators and makes LLM-Judge Accuracy directly comparable across systems.

\begin{table*}[h]
\centering
\scriptsize
\setlength{\tabcolsep}{3pt}
\renewcommand{\arraystretch}{0.92}
\caption{The F1 and LLM-Judge Accuracy (Acc.) for each question category on \textbf{LoCoMo}.}
\label{tab:locomo_main}
\resizebox{0.92\textwidth}{!}{
\begin{tabular}{clcccccccccc}
\toprule
Model & Method
& \multicolumn{2}{c}{Single Hop}
& \multicolumn{2}{c}{Multi Hop}
& \multicolumn{2}{c}{Temporal}
& \multicolumn{2}{c}{Open Domain}
& \multicolumn{2}{c}{Overall} \\
\cmidrule(lr){3-4}
\cmidrule(lr){5-6}
\cmidrule(lr){7-8}
\cmidrule(lr){9-10}
\cmidrule(lr){11-12}
&
& F1 & Acc.
& F1 & Acc.
& F1 & Acc.
& F1 & Acc.
& F1 & Acc. \\
\midrule

\multirow{7}{*}{\rotatebox[origin=c]{90}{GPT-4o-mini}}
& MemoryOS         & 42.95 & 63.85 & 35.18 & 56.74 & 44.41 & 57.63 & 24.63 & 45.83 & 40.69 & 60.13 \\
& Mem0             & 45.01 & 67.30 & 34.64 & 54.96 & 44.36 & 53.27 & 24.89 & 52.08 & 41.72 & 61.17 \\
& MemTree          & 44.90 & 72.77 & 35.66 & 60.64 & 45.11 & 62.30 & 25.43 & 52.08 & 42.04 & 67.08 \\
& MemOS            & 46.95 & 82.52 & 36.95 & 74.47 & 55.24 & 72.90 & 26.79 & 62.50 & 45.59 & 77.79 \\
& Zep              & 46.81 & 83.23 & 33.57 & 69.15 & 55.93 & 70.09 & 24.26 & 61.46 & 44.88 & 76.56 \\
& EverMemOS        & \underline{48.72} & \underline{89.06} & \underline{39.03} & \underline{85.11} & \underline{58.79} & \textbf{82.87} & \underline{28.11} & \textbf{64.58} & \underline{47.76} & \underline{85.52} \\
\cmidrule(lr){2-12}
& \method{} & \textbf{60.20} & \textbf{95.01} & \textbf{47.50} & \textbf{89.01} & \textbf{61.00} & \underline{80.06} & \textbf{28.40} & \underline{63.54} & \textbf{56.06} & \textbf{88.83} \\

\midrule

\multirow{8}{*}{\rotatebox[origin=c]{90}{GPT-4.1-mini}}
& MemoryOS         & 47.20 & 68.49 & 39.80 & 62.06 & 50.30 & 62.62 & 29.40 & 51.04 & 45.38 & 65.00 \\
& Mem0             & 49.08 & 72.53 & 38.77 & 60.28 & 49.22 & 57.63 & 29.78 & 57.29 & 46.02 & 66.23 \\
& MemTree          & 48.35 & 78.36 & 40.25 & 65.25 & 50.90 & 67.29 & 30.10 & 58.33 & 46.26 & 72.40 \\
& MemOS            & 52.16 & 87.63 & 41.36 & 79.79 & 61.58 & 77.88 & 30.65 & 68.75 & 50.80 & 82.99 \\
& Zep              & 51.76 & 88.47 & 38.22 & 74.11 & 57.78 & 75.39 & 28.17 & 67.71 & 49.06 & 81.81 \\
& EverMemOS        & 53.79 & 94.29 & 45.22 & \underline{90.43} & 63.26 & 88.16 & \underline{33.66} & \underline{69.79} & 52.94 & 90.78 \\
\cmidrule(lr){2-12}
& \method{}         & \textbf{57.67} & \underline{96.08} & \textbf{46.81} & 89.01 & \underline{64.00} & \underline{89.72} & \textbf{34.90} & \textbf{72.92} & \textbf{55.58} & \underline{92.01} \\
& \method{} (k=30)  & \underline{56.80} & \textbf{96.31} & \underline{46.79} & \textbf{93.62} & \textbf{64.41} & \textbf{90.03} & 30.62 & \underline{69.79} & \underline{54.92} & \textbf{92.86} \\

\bottomrule
\end{tabular}
}
\end{table*}

\subsection{Main Results}
\label{sec:main_results}

We report F1 and LLM-Judge Accuracy on LoCoMo, LongMemEvalS, and REALTALK in Tables~\ref{tab:locomo_main}, \ref{tab:longmemevals_main}, and \ref{tab:realtalk_main}, respectively.
Overall, \method{} consistently outperforms the baselines, with the largest gains appearing on multi-hop and temporal questions.

On \textbf{LoCoMo}, \method{} achieves the best overall F1 under both backbone LLMs and remains competitive in LLM-Judge Accuracy.
The improvements on multi-hop and temporal categories indicate that graph-based expansion and temporal tree retrieval help locate dispersed and time-sensitive evidence.
On \textbf{LongMemEvalS}, \method{} shows stronger gains, especially on multi-session reasoning, temporal reasoning, and knowledge update, suggesting that the hybrid memory structure is effective for long histories and updated information.
On \textbf{REALTALK}, \method{} also obtains the best overall performance, showing that the proposed retrieval mechanism remains useful under noisier real-world conversations.
These results demonstrate that combining temporal memory abstraction with entity-centered graph retrieval improves long-term memory QA.

\begin{table*}[h]
\centering
\scriptsize
\setlength{\tabcolsep}{4pt}
\caption{Overall results on \textbf{LongMemEvalS}, 
where SSU, MS, SSP, TR, KU, and SSA are the abbreviations of single-session-user, multi-session, single-session-preference, temporal reasoning, knowledge update, and single-session-assistant, respectively.}
\label{tab:longmemevals_main}
\resizebox{\textwidth}{!}{
\begin{tabular}{lcccccccccccccc}
\toprule
Method
& \multicolumn{2}{c}{SSU}
& \multicolumn{2}{c}{MS}
& \multicolumn{2}{c}{SSP}
& \multicolumn{2}{c}{TR}
& \multicolumn{2}{c}{KU}
& \multicolumn{2}{c}{SSA}
& \multicolumn{2}{c}{Overall} \\
\cmidrule(lr){2-3}
\cmidrule(lr){4-5}
\cmidrule(lr){6-7}
\cmidrule(lr){8-9}
\cmidrule(lr){10-11}
\cmidrule(lr){12-13}
\cmidrule(lr){14-15}
& F1 & Acc.
& F1 & Acc.
& F1 & Acc.
& F1 & Acc.
& F1 & Acc.
& F1 & Acc.
& F1 & Acc. \\
\midrule
MemoryOS   & 67.88 & 88.57 & 30.42 & 61.65 & 11.21 & 76.67 & 33.06 & 66.17 & 44.83 & 69.23 & 60.74 & 76.79 & 40.86 & 70.40 \\
Mem0       & 61.73 & 82.86 & 29.86 & 63.16 & 11.58 & 90.00 & 33.47 & 72.18 & 46.21 & 66.67 & 40.36 & 26.79 & 37.91 & 66.40 \\
MemTree    & 69.84 & 87.14 & 34.18 & 63.91 & 10.96 & 63.33 & 36.92 & 69.17 & 48.63 & 69.23 & 68.74 & 75.00 & 44.63 & 70.60 \\
MemOS      & 69.68 & 95.71 & 33.21 & 70.68 & \underline{12.74} & \textbf{96.67} & 34.02 & 76.69 & 45.86 & 79.49 & 58.35 & 67.86 & 42.09 & 78.40 \\
Zep        & 64.11 & 88.57 & 24.83 & 45.11 & 10.26 & 50.00 & 30.72 & 51.88 & 47.09 & 74.36 & 62.35 & 75.00 & 38.70 & 61.20 \\
EverMemOS  & \underline{78.03} & \underline{97.14} & \underline{43.27} & \underline{73.68} & \textbf{13.52} & \underline{93.33} & \underline{43.88} & \underline{77.44} & \underline{58.61} & \underline{89.74} & \underline{79.46} & \underline{83.93} & \underline{52.96} & \underline{82.80} \\
\midrule
\method{} & \textbf{78.70} & \textbf{98.57} & \textbf{46.70} & \textbf{83.46} & 12.08 & \underline{93.33} & \textbf{55.35} & \textbf{84.96} & \textbf{66.60} & \textbf{91.03} & \textbf{82.35} & \textbf{96.43} & \textbf{58.50} & \textbf{89.20} \\
\bottomrule
\end{tabular}
}
\end{table*}

\begin{table*}[h]
\centering
\scriptsize
\setlength{\tabcolsep}{3pt}
\renewcommand{\arraystretch}{0.90}
\caption{Overall results on \textbf{REALTALK}.}
\label{tab:realtalk_main}
\resizebox{0.65\textwidth}{!}{
\begin{tabular}{l*{8}{c}}
\toprule
\multirow{2}{*}{Method}
& \multicolumn{2}{c}{Multi-hop}
& \multicolumn{2}{c}{Temporal}
& \multicolumn{2}{c}{Open-domain}
& \multicolumn{2}{c}{Overall} \\
\cmidrule(lr){2-3}
\cmidrule(lr){4-5}
\cmidrule(lr){6-7}
\cmidrule(lr){8-9}
& F1 & Acc.
& F1 & Acc.
& F1 & Acc.
& F1 & Acc. \\
\midrule
MemoryOS   & 26.83 & 51.16 & 13.02 & 36.68 & 22.50 & 62.04 & 20.14 & 46.43 \\
Mem0       & 28.44 & 53.49 & 14.51 & 37.62 & 23.37 & 63.89 & 21.58 & 48.08 \\
MemTree    & 31.18 & 66.45 & 25.67 & 58.93 & \underline{25.24} & 58.33 & 27.88 & 61.95 \\
MemOS      & 30.66 & 65.12 & 25.19 & 56.74 & 24.46 & 58.33 & 27.34 & 60.44 \\
Zep        & 28.15 & 62.79 & 23.18 & 54.23 & 20.54 & 54.63 & 24.84 & 57.83 \\
EverMemOS  & \underline{35.78} & \underline{72.76} & \underline{41.77} & \textbf{80.56} & 25.03 & \underline{71.30} & \underline{36.81} & \underline{75.96} \\
\midrule
\method{}  & \textbf{37.41} & \textbf{82.39} & \textbf{45.59} & \underline{74.29} & \textbf{26.03} & \textbf{77.78} & \textbf{39.31} & \textbf{78.16} \\
\bottomrule
\end{tabular}
}
\end{table*}

\begin{figure*}[h]
    \centering
    \includegraphics[width=1.0\textwidth]{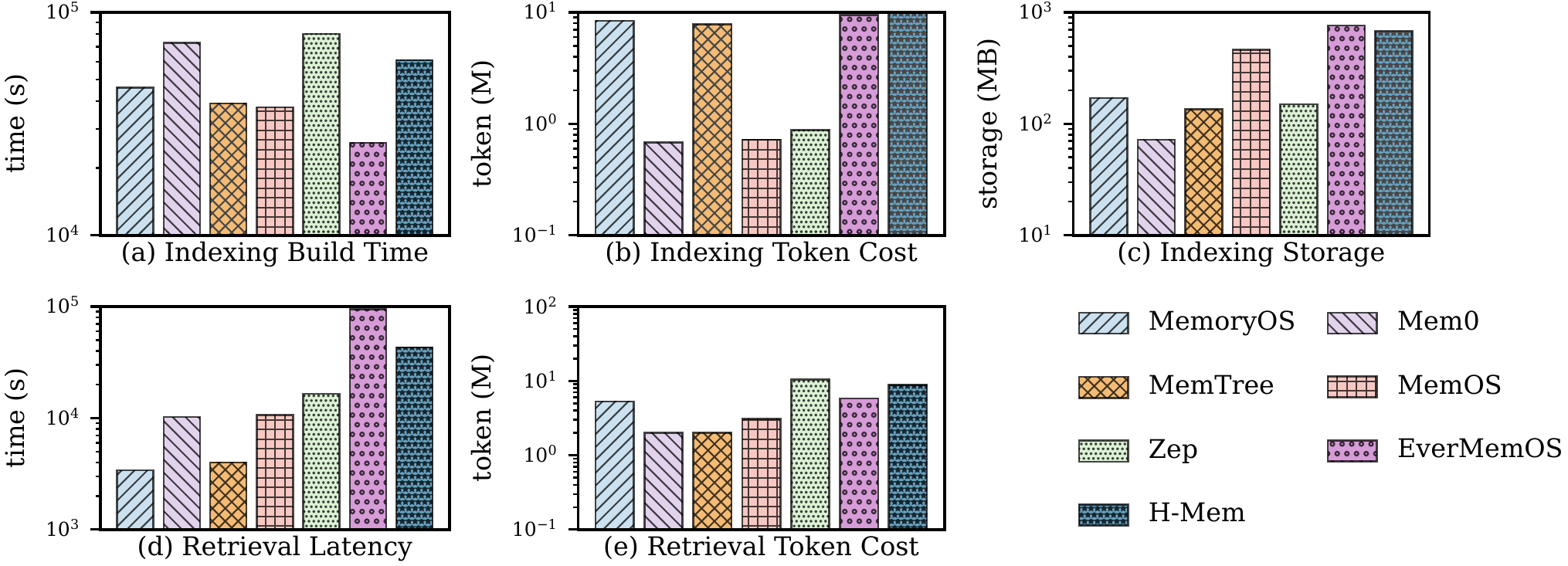}
    \caption{Cumulative indexing and retrieval costs of \method{} and baselines on \textbf{LoCoMo}, including indexing time, indexing token cost, index storage, retrieval latency, and retrieval token cost.}
    \label{fig:efficiency_breakdown}
\end{figure*}

Besides QA accuracy, we compare \method{} with baselines from different aspects in both offline indexing and online retrieval stages.
For offline indexing, we use three metrics: indexing time, indexing token cost, and index storage, where the former two measure the time and token cost for building the index respectively, and the latter one reports the final index size.
As shown in Figure~\ref{fig:efficiency_breakdown}(a), \method{} has a moderate indexing time.
This indicates that constructing the hybrid structure does not make the offline indexing process prohibitively slow.
Figures~\ref{fig:efficiency_breakdown}(b)--(c) show that \method{} has relatively high indexing token cost and index storage. Its indexing token cost is close to the high-cost group, including
MemoryOS, MemTree, and EverMemOS. Its index
storage is higher than most baselines, while still lower than EverMemOS.
This is expected because \method{} needs to construct memory events, memory summaries, entities, relations, and entity profiles, while maintaining both the temporal and semantic tree and the knowledge graph. Therefore, the additional offline cost is consistent with the goal of modeling memory evolution and supporting multi-hop retrieval.

For online retrieval, we report retrieval latency and retrieval token cost, which measure the latency and token cost during retrieval, respectively.
As shown in Figure~\ref{fig:efficiency_breakdown}(d), \method{} remains lower than EverMemOS. This overhead mainly comes from query decomposition, graph exploration, bottom-up tree search, and evidence
re-ranking.
In terms of retrieval token cost, Figure~\ref{fig:efficiency_breakdown}(e) shows that \method{} is higher than most baselines but lower than Zep.
This is mainly because \method{} retrieves evidence from both the tree and the graph, while deduplication and re-ranking help control the final evidence chain.
Overall, \method{} introduces additional indexing and retrieval costs compared with several simpler memory mechanisms, but these costs remain reasonable considering its more expressive hybrid memory structure.

\subsection{Ablation Study}
\label{sec:ablation}
We perform ablation studies to identify the contribution of key components in \method{}.
Specifically, \textit{w/o tree} disables the tree and removes bottom-up tree search;
\textit{w/o graph} disables entity nodes, relations, and multi-hop entity expansion;
\textit{w/o long-term memory} removes upper-level memory summaries and only searches memory events and original memory fragments;
\textit{w/o memory robustness} removes the robustness term from evidence scoring;
\textit{w/o missing-info query} disables follow-up retrieval when the first-pass evidence is insufficient;
and \textit{w/o entity profile} removes entity profiles.

As shown in Table~\ref{tab:ablation_main}, removing the tree causes the largest performance drop among all ablation variants.
This shows that the tree structure is the most critical component of \method{}, since it organizes memory data across different time windows and semantic granularities and supports bottom-up
retrieval.
The second largest drop comes from removing the graph.
This confirms that entity-centered information and multi-hop relationships are important complements to the tree.
The variant without long-term memory also shows a clear performance drop, indicating that memory summaries are useful for preserving abstract and persistent information at retrieval.
The variants without memory robustness, missing-information query, and entity profile also show consistent but smaller drops.
This indicates that repeated memory reinforcement, follow-up retrieval for insufficient first-pass evidence, and entity-centered profiles all provide useful support for improving memory retrieval.

Overall, the ablation results show that the performance gains of \method{} mainly come from the cooperation between the temporal and semantic tree and the entity-centered knowledge graph.

\begin{table}[h]
\centering
\small
\setlength{\tabcolsep}{6pt}
\caption{Ablation study on \textbf{LoCoMo}.}
\label{tab:ablation_main}
\begin{tabular}{lcc}
\toprule
Variant & F1 & Acc. \\
\midrule
\method{}(Full)                 & 55.58 & 92.01 \\
w/o tree                       & 49.76 & 82.73 \\
w/o graph                      & 52.20 & 87.86 \\
w/o long memory         & 53.39 & 89.48 \\
w/o memory robustness                  & 53.72 & 89.09 \\
w/o missing-info query            & 54.82 & 90.97  \\
w/o entity profile             & 54.05 & 90.32  \\
\bottomrule
\end{tabular}
\end{table}


\section{Conclusion}
\label{sec:conclude}

In this work, we present \method{}, a novel memory mechanism for evolving and retrieving agent memory via a hybrid structure.
Particularly, \method{} builds a temporal and semantic tree to organize memory data  across different time windows and semantic granularities, where short-term memory data can evolve progressively into long-term memory data.
It also constructs a knowledge graph to capture entity-centered information and multi-hop relations.
To support the QA task, \method{} retrieves relevant evidence by exploiting the above hybrid structure. 
Our method achieves state-of-the-art performance on three public long-term memory benchmarks, demonstrating its superiority over existing baselines for the QA task.
In the future, we will further improve \method{} to support multimodal memory data and explore its deployment in real-world agent applications.

\clearpage
\bibliographystyle{unsrtnat}
\bibliography{references}

\clearpage
\onecolumn
\raggedbottom
\appendix
\section{Experiment Details}
\label{app:experiment_details}

\subsection{Dataset and Index Statistics}
\label{app:dataset_index}

This subsection provides supplementary corpus statistics and index statistics for the benchmarks used in the main paper.
Table~\ref{tab:dataset_details} summarizes the basic properties of the three datasets, including dataset scale, average conversation length, construction setting, and data source.

\begin{table}[H]
\centering
\small
\setlength{\tabcolsep}{6pt}
\renewcommand{\arraystretch}{1.15}
\caption{Dataset details.}
\label{tab:dataset_details}
\begin{tabular}{lccc}
\toprule
Metric & LoCoMo & LongMemEvalS & REALTALK \\
\midrule
\# Conv. & 10 & 500 & 10 \\
\# Questions & 1540 & 500 & 728 \\
Avg. Tokens / Conv. & 16.6K & 115.0K & 20.7K \\
Setting & LLM-simulated & LLM-simulated & Crowdsourced \\
Source & LLM-gen. + crowdsourcing & synthetic user--assistant chats & messaging app dialogues \\
\bottomrule
\end{tabular}
\end{table}

The three datasets cover complementary evaluation settings.
LoCoMo provides dense question supervision over long multi-session dialogues, while LongMemEvalS contains substantially longer conversational histories and is therefore more challenging for long-context memory retrieval.
REALTALK further complements the two LLM-simulated benchmarks with real-world human--human conversations, making it useful for evaluating robustness under noisier conversational conditions.

Table~\ref{tab:index_stats} reports the average index statistics after offline indexing, including raw conversation size, fragment count, hierarchical event count, and entity-graph statistics.

\begin{table}[H]
\centering
\small
\setlength{\tabcolsep}{6pt}
\renewcommand{\arraystretch}{1.15}
\caption{Average index statistics per conversation after offline indexing.}
\label{tab:index_stats}
\begin{tabular}{lccc}
\toprule
Metric & LoCoMo & LongMemEvalS & REALTALK \\
\midrule
Avg. Raw Conv Size (MB) & 7.9 & 12.7 & 9.7 \\
Avg. Fragments & 596.5 & 493.5 & 894.4 \\
Avg. L1 Events & 995.1 & 1819.6 & 1209.7 \\
Avg. L2 Events & 563.7 & 1008.7 & 840.1 \\
Avg. L3 Events & 419.7 & 594.6 & 730.1 \\
Avg. L4 Events & 100.9 & 325.8 & 202.7 \\
Avg. Entities & 828.6 & 3267.6 & 1013.3 \\
Avg. Entity Edges & 3670.8 & 15911.5 & 4556.8 \\
\bottomrule
\end{tabular}
\end{table}

The index statistics show that different benchmarks stress different aspects of long-term memory.
LongMemEvalS produces the largest event hierarchy and entity graph, reflecting its longer conversational histories and more complex cross-session dependencies.
REALTALK has the largest number of fragments on average, which is consistent with the greater heterogeneity and noisiness of real-world conversations.
Across all datasets, \method{} expands raw fragments into multi-level events and entity graphs, enabling retrieval at both temporal and entity-centric granularities.

\subsection{Implementation Details}
\label{app:implementation}

{\bf Model configuration.}
We evaluate \method{} with GPT-4o-mini~\citep{openai2024gpt4omini}
and GPT-4.1-mini~\citep{openai2025gpt41} as backbone LLMs.
For semantic retrieval and evidence reranking, the default configuration uses Qwen3-Embedding-4B and Qwen3-Reranker-4B~\citep{zhang2025qwen3embedding}.
We also evaluate a lighter configuration with Qwen3-Embedding-0.6B and
Qwen3-Reranker-0.6B~\citep{zhang2025qwen3embedding} in the hyperparameter analysis.
All experiments are conducted on a Linux server equipped with an Intel Xeon 2.0GHz CPU, 1024GB of memory, and 8 NVIDIA GeForce RTX A5000 GPUs, each with 24GB of VRAM.

{\bf Baseline configuration.}
For each baseline, we follow its original memory organization and retrieval design as closely as possible.
To reduce confounding factors, all baselines are evaluated under the same experimental framework.
When semantic retrieval or reranking is required, all methods use the same embedding and reranking configurations.
All methods are evaluated with the same answer simplification, normalization, F1 computation, and LLM-Judge protocol.

{\bf Tree Index Hyperparameter.}
For the temporal-semantic tree, we use four levels corresponding to day-, week-, month-, and year-level memory organization.
The leaf level stores fine-grained memory events, while upper levels store consolidated summaries over longer temporal windows.
Specifically, $\beta_l$ denotes the temporal window size at level $l$, and $\alpha_l$ denotes the semantic clustering threshold for consolidation within the corresponding temporal window.
By default, we use day, week, month, and year as the temporal windows from L1 to L4, and set the consolidation thresholds of L2, L3, and L4 to $0.8$, $0.7$, and $0.6$, respectively.
The threshold gradually decreases at higher levels because higher-level summaries are expected to capture more abstract and persistent memory patterns.
We also tested two alternative threshold schedules: a conservative schedule $(0.9,0.8,0.7)$ and an aggressive schedule $(0.7,0.6,0.5)$.
The conservative schedule retains more fine-grained nodes but makes the upper-level index more fragmented, while the aggressive schedule yields a more compact index but risks over-consolidating heterogeneous memories.
To avoid unnecessary high-level consolidation when the memory history is short, the maximum active level is determined by the memory age: histories shorter than 7 days activate day and week levels; histories between 7 and 30 days activate day, week, and month levels; otherwise, all four levels are activated.

{\bf Memory robustness and scoring hyperparameter.}
For evidence ranking, \method{} combines semantic similarity, temporal alignment, and memory robustness.
The event-level relevance score is computed as
\[
s(m,q)=
w_{\mathrm{sem}}\cdot \mathrm{sim}(m,q)
+
w_{\mathrm{time}}\cdot \mathrm{time}(m,q)
+
w_{\mathrm{mem}}\cdot R(m,t),
\]
where $\mathrm{sim}(m,q)$ is the semantic similarity between memory $m$ and query $q$, $\mathrm{time}(m,q)$ measures temporal alignment when an explicit temporal hint is available, and $R(m,t)$ denotes the memory robustness score.
We set $w_{\mathrm{sem}}=0.70$, $w_{\mathrm{time}}=0.15$, and $w_{\mathrm{mem}}=0.15$ by default.
When no explicit temporal hint is available, the temporal alignment term is set to zero.

The memory robustness score follows an Ebbinghaus-style decay with reinforcement:
\[
R(m,t)=
\exp\left(
-\frac{t-r_m}
{\tau(1+\eta\log(1+n_m))}
\right),
\]
where $r_m$ is the latest reinforcement timestamp of memory $m$, $n_m$ is the number of reinforcements, $\eta$ controls the reinforcement effect, and $\tau$ is the decay time scale.
We set $\tau=365$ days and $\eta=0.5$ by default.
In implementation, both $t-r_m$ and $\tau$ are converted to seconds. 
Thus, when $\tau=365$ days, an unreinforced memory after one year has
\[
R=\exp(-1)\approx 36.8\%,
\]
corresponding to a decay of about $63.2\%$.
This choice is reasonable for long-term agent memory because many user preferences, relationships, and recurring facts should remain retrievable over a year-scale horizon rather than being rapidly forgotten.
At the same time, the robustness term is only a weak ranking prior with $w_{\mathrm{mem}}=0.15$, so it does not dominate semantic relevance or explicit temporal matching.
Repeatedly reinforced memories decay more slowly through the factor $1+\eta\log(1+n_m)$, which allows stable and recurring memories to remain more salient during retrieval.

{\bf Entity extraction and disambiguation.}
For graph construction, \method{} processes each memory fragment through entity and relation extraction, entity resolution, and relation insertion.
Specifically, \method{} uses an LLM-based information extraction prompt to extract entities and relations from each memory fragment.
Each extracted entity is represented as a structured record containing its surface name, entity type, optional text span, role, salience score, and auxiliary metadata.
The entity type is normalized into a predefined type set, including person, organization, location, event, product, work, date, time, and other.
Each extracted relation contains a source entity, a target entity, a relation label, a confidence score, and an optional text span.
If LLM-based entity extraction fails or returns no valid entity, \method{} falls back to spaCy NER to obtain a best-effort entity set.

After extraction, the extracted entities are normalized and resolved into entity nodes.
The normalization step lower-cases entity names, removes unnecessary punctuation, normalizes whitespace, maps entity types into the predefined type set, and optionally uses lemmatization.
For each newly extracted entity, \method{} first checks whether it exactly matches an existing entity node with a compatible entity type.
If no exact match is found, the entity is compared with existing entity names and aliases using token overlap and fuzzy string matching.
If the entity satisfies the matching criteria, it is merged into the existing entity node, and its surface name is stored as an alias when applicable.
Otherwise, a new entity node is created.
Each resolved entity node is then linked to the original memory fragment containing it, preserving provenance for later evidence verification.

We distinguish entity merging from graph repair.
Entity merging indicates that two extracted entities are resolved as the same real-world entity and therefore share one entity node.
Graph repair does not merge nodes or modify aliases.
Instead, for short single-token names or nickname-like variants that remain as separate nodes, \method{} may add an \textit{overlap} edge based on prefix/suffix matching.
This edge is used only to improve graph traversal recall during retrieval and does not indicate identity equivalence.

Finally, the extracted relations are mapped to resolved entity nodes before being inserted into the graph.
That is, the source and target entities of each relation are first resolved to their corresponding entity nodes.
The relation is then inserted as an entity--entity edge with its relation label, confidence weight, timestamp, and supporting evidence.
If the same relation between the same resolved entities already exists, \method{} merges the supporting evidence instead of creating a duplicate edge.
Since the current implementation does not rely on calibrated extraction confidence for pruning, it uses conservative type-compatible matching, bounded fuzzy matching, and provenance links to reduce the impact of extraction noise.
During retrieval, the provenance links allow entity and relation evidence to be verified against the original memory fragments before answer generation.

\subsection{Evaluation Details}
\label{app:evaluation_details}

We provide the implementation details for the evaluation protocol used in the main paper.

{\bf Answer Simplification.}
Before lexical evaluation, the predicted answer is simplified into a short factual form.
This step reduces verbosity artifacts in generative answers and makes token-overlap metrics more faithful to semantic correctness.
Answer simplification is only used for lexical F1 evaluation and does not affect the generated answer used by the LLM judge.
The full answer simplification prompt is provided in Appendix~\ref{app:evaluation_prompts}.

{\bf Normalization and F1 Computation.}
After simplification, both the prediction and the reference answer are normalized before F1 computation.
The normalization pipeline includes Unicode normalization, lower-casing, punctuation and article removal, and whitespace collapsing.
Let $\hat{A}$ denote the normalized prediction and $A$ denote the normalized reference answer.
After tokenization, we compute token-level precision, recall, and F1 as
\[
P = \frac{|\mathrm{Tok}(\hat{A}) \cap \mathrm{Tok}(A)|}{|\mathrm{Tok}(\hat{A})|},
\quad
R = \frac{|\mathrm{Tok}(\hat{A}) \cap \mathrm{Tok}(A)|}{|\mathrm{Tok}(A)|},
\quad
\mathrm{F1} = \frac{2PR}{P+R}.
\]
where $\mathrm{Tok}(\cdot)$ denotes the token multiset after normalization, and $\cap$ denotes multiset intersection.
If both precision and recall are zero, the F1 score is set to zero.

{\bf LLM-Judge Protocol.}
For answer evaluation, we follow the LLM-as-a-judge prompt used in \textsc{MemGPT}~\citep{packer2024memgpt}, where the judge is given the question, the gold answer, and the predicted answer, and returns \texttt{CORRECT} or \texttt{WRONG} under a semantically tolerant criterion.
The same judging prompt is used for all methods to ensure direct comparability.
The full LLM-Judge prompt is provided in Appendix~\ref{app:evaluation_prompts}.

\subsection{Additional Experiments}
\label{app:additional_experiments}

We provide additional analyses of \method{} on retrieval hyperparameters and retrieval planning behaviors, including the retriever scale, memory-scope distribution, and missing-information query behavior.

\begin{figure}[H]
\centering
\includegraphics[width=0.55\linewidth]{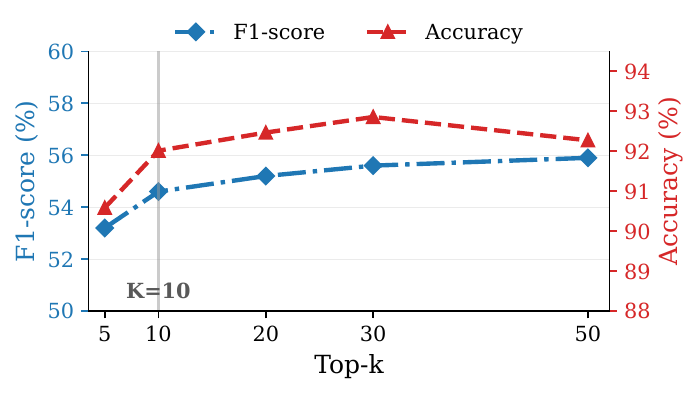}
\caption{Sensitivity to the top-$k$ retrieval budget on \textbf{LoCoMo}.}
\label{fig:topk_figure}
\end{figure}
\noindent$\bullet$ {\bf Effect of top-$k$.}
We analyze the sensitivity of \method{} to the top-$k$ retrieval budget on \textbf{LoCoMo}.
In our implementation, $k$ controls the budget for entity-related fragment retrieval and memory event retrieval.
Figure~\ref{fig:topk_figure} reports the F1 and LLM-Judge Accuracy under different top-$k$ settings.
Overall, \method{} remains stable across a moderate range of $k$.
Increasing $k$ provides a larger candidate pool and improves F1, suggesting that more candidate evidence helps recover useful supporting information.
Accuracy also improves as $k$ increases from 5 to 30, but slightly decreases at $k=50$, indicating that overly large candidate pools may introduce redundant or noisy context.
Therefore, we use the setting that balances F1 and LLM-Judge Accuracy as the default configuration in the main experiments.

\noindent$\bullet$ {\bf Effect of retrieval model components.}
Table~\ref{tab:hparam_models} compares different retrieval model configurations, including an embedding-only setting, a lighter 0.6B embedding/reranker pair, and a stronger 4B embedding/reranker pair.
Compared with the embedding-only setting, adding the reranker improves both F1 and LLM-Judge Accuracy, showing that reranking helps prioritize more useful evidence for downstream reasoning.
Notably, the embedding-only setting also performs well.
This is because \method{} retrieves over clean and atomic memory events rather than raw conversational chunks.
These extracted events contain less irrelevant dialogue context and have clearer semantic boundaries, which makes embedding similarity more reliable for matching queries to candidate evidence.
The stronger 4B configuration further achieves slightly higher F1 and accuracy than the 0.6B configuration, while the lighter model pair remains competitive.
This indicates that \method{} benefits from stronger retrieval models, but its performance is mainly supported by the proposed memory structure rather than by retrieval model scale alone.
\begin{table}[H]
\centering
\small
\setlength{\tabcolsep}{6pt}
\renewcommand{\arraystretch}{1.12}
\caption{Sensitivity to embedding and reranker models on \textbf{LoCoMo}.}
\label{tab:hparam_models}
\begin{tabular}{lcc}
\toprule
Retrieval Model & F1 & Acc. \\
\midrule
Qwen3-Embedding-0.6B + Qwen3-Reranker-0.6B & 54.96 & 91.62 \\
Qwen3-Embedding-4B + Qwen3-Reranker-4B & 55.58 & 92.01 \\
Qwen3-Embedding-4B only & 53.72 & 89.81 \\
\bottomrule
\end{tabular}
\end{table}

\begin{figure}[H]
\centering
\includegraphics[width=0.68\linewidth]{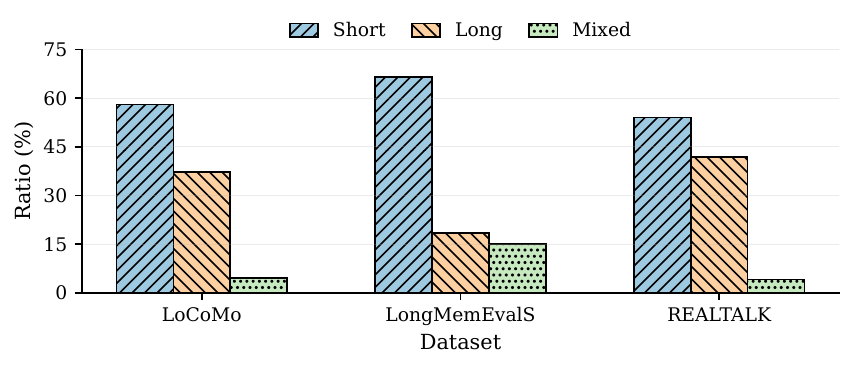}
\caption{Distribution of memory scopes predicted by the retrieval planner.}
\label{fig:memory_scope_distribution}
\end{figure}

$\bullet$ {\bf Memory scope distribution.}
We analyze the memory scopes predicted by the retrieval planner.
For each sub-query, the planner selects one scope from \textsc{Short}, \textsc{Long}, and \textsc{Mixed}, which determines the memory levels used during tree-based retrieval.
\textsc{Short} is mainly used for moment-specific evidence, such as concrete events, recent actions, or temporally localized facts.
\textsc{Long} is used when the query requires stable or persistent memory, such as long-term preferences, relationships, or recurring facts.
\textsc{Mixed} is selected when both fine-grained situational evidence and higher-level memory summaries are needed.
Figure~\ref{fig:memory_scope_distribution} reports the distribution of predicted memory scopes.

The distribution shows that the planner does not rely on a fixed retrieval granularity.
Instead, it adaptively selects the retrieval scope according to the information need of each sub-query.
Across the three datasets, \textsc{Short} remains the most frequently selected scope, while \textsc{Long} also accounts for a substantial portion of sub-queries, especially on LoCoMo and REALTALK.
This indicates that long-term conversational QA requires both moment-specific evidence and stable reusable memory.
We further compare the adaptive scope strategy with a fixed \textsc{Mixed} retrieval policy, where every sub-query retrieves both fine-grained events and higher-level summaries.
The fixed \textsc{Mixed} policy also achieves comparable QA performance, with an accuracy of $92.21\%$.
This shows that retrieving both fine-grained events and higher-level summaries can cover most required evidence.
However, fixed \textsc{Mixed} uses a much larger evidence context and incurs about $1.8\times$ retrieval token cost compared with the planner-based strategy.
Therefore, the main advantage of memory scope prediction is retrieval efficiency: it preserves comparable answer quality while reducing token usage by adaptively selecting \textsc{Short}, \textsc{Long}, or \textsc{Mixed} for each sub-query.

$\bullet$ {\bf Missing-information query analysis.}
We further analyze how often \method{} triggers missing-information queries.
A missing-information query is generated only when the first-pass evidence is insufficient to answer a sub-query with the required specificity.
Typical cases include unresolved entities, vague references, missing temporal anchors, underspecified event descriptions, and multiple plausible candidates.
Table~\ref{tab:missing_info_ratio} reports the triggering statistics of missing-information queries.

\begin{table}[H]
\centering
\small
\setlength{\tabcolsep}{6pt}
\renewcommand{\arraystretch}{1.12}
\caption{Triggering statistics of missing-information queries.}
\label{tab:missing_info_ratio}
\begin{tabular}{lcccc}
\toprule
Dataset & Questions & Total Sub-queries & Missing-info Sub-queries & Trigger Ratio \\
\midrule
LoCoMo & 1540 & 1777 & 113 & 6.36\% \\
LongMemEvalS & 500 & 792 & 120 & 15.15\% \\
REALTALK & 728 & 930 & 92 & 9.89\% \\
\bottomrule
\end{tabular}
\end{table}

Instead of simply increasing the retrieval depth for all queries, \method{} performs targeted follow-up retrieval only when an evidence gap is detected.
The generated missing-information query is required to ask for the missing slot rather than paraphrasing the original sub-query, and it must contain at least one concrete anchor from the first-pass evidence.
This strategy helps recover missing bridge evidence while avoiding unnecessary retrieval cost for queries that can already be answered from the first-pass evidence.

\subsection{Controlled Stability Analysis}
\label{app:stability}

To examine the stability of \method{}, we conduct three controlled repeated evaluations on representative main settings, including LoCoMo, LongMemEvalS, and REALTALK. Since \method{} is a retrieval-based memory system and does not train a task-specific neural model, these repeated evaluations are used to measure evaluation-pipeline stability rather than training-run variability. To reduce accidental variation introduced by the planning stage, we reuse the same planner outputs across the three runs, including the decomposed sub-queries and predicted memory scopes. In addition, all LLM-based components are evaluated with temperature set to 0.

For each setting, we report the LLM-Judge Accuracy and F1 score of each run, together with the mean and sample standard deviation across the three runs. The reported error range corresponds to $\pm 1$ sample standard deviation.

\begin{table*}[t]
\centering
\scriptsize
\setlength{\tabcolsep}{4pt}
\renewcommand{\arraystretch}{1.05}
\caption{Controlled stability analysis over three repeated evaluations. We reuse the same planner outputs across runs and set the temperature of all LLM-based components to 0. We report the F1 score and LLM-Judge Accuracy of each run, together with the mean and sample standard deviation. The error range corresponds to $\pm 1$ sample standard deviation.}
\label{tab:error_bar_stats}
\resizebox{0.95\textwidth}{!}{
\begin{tabular}{lcccccccc}
\toprule
\multirow{2}{*}{Dataset} 
& \multicolumn{4}{c}{F1}
& \multicolumn{4}{c}{LLM-Judge Accuracy} \\
\cmidrule(lr){2-5}
\cmidrule(lr){6-9}
& Run 1 & Run 2 & Run 3 & Mean $\pm$ Std.
& Run 1 & Run 2 & Run 3 & Mean $\pm$ Std. \\
\midrule

LoCoMo 
& 55.58 & 55.39 & 55.66 & $55.54 \pm 0.14$
& 92.01 & 91.69 & 92.14 & $91.95 \pm 0.23$ \\

LongMemEvalS
& 58.50 & 58.37 & 57.71 & $58.19 \pm 0.42$
& 89.20 & 89.00 & 88.00 & $88.73 \pm 0.64$ \\

REALTALK
& 39.31 & 39.10 & 38.69 & $39.03 \pm 0.31$
& 78.16 & 77.74 & 76.92 & $77.61 \pm 0.63$ \\

\bottomrule
\end{tabular}
}
\end{table*}

The results show that \method{} remains stable across repeated evaluations. The standard deviations are small for both LLM-Judge Accuracy and F1, indicating that the reported improvements are not mainly caused by incidental evaluation fluctuations. This is consistent with the controlled evaluation setup, where the planner outputs are fixed and deterministic decoding is used for all LLM-based components.

\subsection{Limitations}
\label{app:limitations}

Although \method{} improves long-term conversational memory retrieval, it still has several limitations.
First, \method{} relies on LLM-based memory construction and retrieval planning.
Therefore, its performance can be affected by imperfect memory-unit extraction, inaccurate query decomposition, and incorrect memory-scope prediction.
Second, the hybrid tree-graph memory structure introduces additional offline indexing cost and storage overhead compared with simpler fragment-level memory systems.
Although the online retrieval latency remains moderate, the indexing stage may become more expensive for extremely long or frequently updated conversations.
Third, our experiments are conducted on existing long-term memory benchmarks.
While these benchmarks cover simulated and real-world conversational settings, further evaluation is needed in deployed interactive agents, longer usage periods, and broader real-world domains.
Finally, long-term memory systems may store and retrieve user-specific information, which introduces potential privacy risks such as unintended retention, sensitive inference, or inappropriate memory reuse.
Practical deployments should therefore include explicit user consent, memory editing and deletion mechanisms, access control, and transparent memory auditing.

\subsection{Assets and Licenses}
\label{app:assets_licenses}

We use existing benchmarks, baselines, and model APIs or publicly available models in our experiments.
The datasets and baselines are cited in the main paper and used only for research evaluation.
For model usage, we follow the corresponding API terms of service or model usage terms.
We do not redistribute any third-party datasets, model weights, or proprietary model outputs as new assets.

\section{Case Studies}
\label{app:case_studies}

We present two representative qualitative examples of \method{}. The first illustrates multi-step sub-query decomposition for a multi-entity question, while the second illustrates missing-information guided retrieval, where the system detects that first-pass evidence is insufficiently specific and issues a bridge-style follow-up query.

\begin{center}

\begin{tcolorbox}[
    enhanced jigsaw,
    breakable,
    width=0.96\textwidth,
    colback=white,
    colframe=black,
    boxrule=0.8pt,
    arc=1pt,
    left=6pt,right=6pt,top=6pt,bottom=6pt,
    title={\textbf{(A) Case Study: Multi-step Sub-query Decomposition}},
    colbacktitle=black,
    coltitle=white,
    fonttitle=\bfseries
]
\small

\textbf{Question.}
What subject have Caroline and Melanie both painted?

\vspace{0.35em}
\hrule
\vspace{0.45em}

\textbf{Planner output.}
\method{} decomposes the question into two entity-specific subqueries:
\begin{itemize}
    \item \textbf{q1}: What subjects has Caroline painted? \hfill (\textsc{SHORT}, \textsc{global})
    \item \textbf{q2}: What subjects has Melanie painted? \hfill (\textsc{SHORT}, \textsc{global})
\end{itemize}

\vspace{0.35em}
\hrule
\vspace{0.45em}

\textbf{Retrieved evidence.}
\begin{itemize}
    \item \textbf{q1}: Caroline shared paintings involving sunsets and floral themes.
    \item \textbf{q2}: Melanie shared paintings including a lake sunrise, a sunset scene, and other nature-inspired subjects.
\end{itemize}

\vspace{0.35em}
\hrule
\vspace{0.45em}

\textbf{Prediction.}
\textbf{sunsets}

\textbf{Gold answer.}
\textbf{Sunsets}

\end{tcolorbox}

\vspace{0.8em}

\begin{tcolorbox}[
    enhanced jigsaw,
    breakable,
    width=0.96\textwidth,
    colback=white,
    colframe=black,
    boxrule=0.8pt,
    arc=1pt,
    left=6pt,right=6pt,top=6pt,bottom=6pt,
    title={\textbf{(B) Case Study: Missing-Information Guided Retrieval}},
    colbacktitle=black,
    coltitle=white,
    fonttitle=\bfseries
]
\small

\textbf{Question.}
Which popular music composer's tunes does Tim enjoy playing on the piano?

\vspace{0.35em}
\hrule
\vspace{0.45em}

\textbf{First-pass evidence.}
The first retrieval pass finds evidence that Tim enjoys playing \emph{a theme from a movie he really likes} on the piano, but the composer is not explicitly named.

\vspace{0.35em}
\hrule
\vspace{0.45em}

\textbf{Reasoner decision.}
The first-pass reasoner marks the subquery as \texttt{missing\_info=true}, because the current evidence is suggestive but still underspecified for answering the question at the required level of specificity.

\vspace{0.35em}
\hrule
\vspace{0.45em}

\textbf{Generated missing-info query.}
\begin{quote}
Which popular music composer created the theme from the movie that Tim enjoys playing on the piano?
\end{quote}

\vspace{0.35em}
\hrule
\vspace{0.45em}

\textbf{Second-pass evidence and answer.}
The follow-up retrieval uncovers the missing bridge between Tim's favorite movie theme and the intended composer, allowing \method{} to produce the final answer:
\textbf{John Williams}

\vspace{0.35em}

\textbf{Gold answer.}
\textbf{John Williams}

\end{tcolorbox}

\captionof{figure}{Two representative qualitative examples of \method{}. (A) \method{} decomposes a multi-entity question into two entity-specific sub-queries and retrieves evidence under global coverage before final synthesis. (B) When first-pass evidence is insufficiently specific, \method{} explicitly detects missing information and generates a bridge-style follow-up query for targeted second-pass retrieval.}
\label{fig:case_studies_combined}

\end{center}

\section{Prompt Templates}
\label{app:prompts}

This appendix provides the core prompt templates used by \method{}.
We include prompts for memory construction, evaluation, retrieval planning, and missing-information query generation.

\subsection{Memory Construction Prompts}
\label{app:memory_construction_prompts}

\begin{promptbox}{Memory extraction prompt}
You are an information extraction engine.
Your job is to extract the maximally fine-grained set of atomic, queryable memory units from conversation fragments.

Return ONLY valid JSON. No markdown. No explanations.

Definition:
- A memory unit is the smallest self-contained piece of information that could answer a future query.
- Memory units include events, facts, plans, preferences, relationships, statuses, goals, challenges, collections/items, and named-entity attributes.
- Each unit must be grounded in the provided fragments via frag_ids.
- Do NOT write narrative summaries or high-level commentary.

Granularity rules:
- Split aggressively: prefer many small units over fewer large ones.
- One unit = one claim.
- If a sentence lists multiple items, create one unit per item.
- If a statement contains multiple attributes, create separate units for each attribute.
- Keep units independently queryable.

Hard requirements:
Output JSON with the exact top-level key: "events".
Each memory unit must include:
- "event_text": a single canonical sentence describing one claim.
- "frag_ids": list of supporting fragment ids.
- "time_range": [start_ts, end_ts] as unix timestamps in seconds.
- "event_type": one of ["episodic","plan","preference","fact","relationship","status","procedure","other"].

Additional rules:
- Preserve named entities, numbers, dates, locations, and constraints.
- Do not invent unsupported details.
- Resolve vague references when the referent is clear from nearby fragments.
- Deduplicate only if two units express the same claim with the same specificity.
- Include short-lived chatter only if it creates a queryable unit.

Cross-fragment change and linkage:
- Consider how fragments change or relate to earlier fragments.
- If a meaningful change is detected, extract an additional unit describing the change.
- Change types include plans, preferences, goals, status, roles, locations, and availability.
- If a later fragment answers, corrects, or refers to an earlier fragment, extract a linkage unit if it is queryable.
- Delta units should use templates such as:
  "Previously <X>; now <Y>." or "Earlier <X>; later <Y>."

Optional fields:
- participants
- location
- confidence
- notes

If no meaningful units can be extracted, return:
{"events": []}
\end{promptbox}

\begin{promptbox}{Memory consolidation prompt}
You are a long-term memory consolidation engine.
Your job is to consolidate lower-level memory units into stable, long-term consolidated events.

Return ONLY valid JSON. No markdown. No explanations.

Key idea:
Consolidation is not summarization.
It converts repeated, stable, or high-signal units into canonical long-term memories.
Do NOT force coverage; many lower-level units may remain unmerged.

Hard constraints:
- Output JSON with top-level keys: "consolidated_events" and "unmerged_event_ids".
- Only output a consolidated event if it is supported by at least two distinct source events.
- Each consolidated event must have source_event_ids length >= 2.
- Do NOT invent facts. Preserve names, numbers, dates, and constraints.
- Each consolidated event should represent one stable memory.
- If the evidence evolves, record the stable core and evolution notes.
- If the evidence conflicts, record the conflict rather than choosing one side.

Merge rules:
- Only merge units that share the same subject/entity and the same relation or attribute.
- Prefer type-homogeneous merges.
- For itemized lists, consolidate into a collection memory only when all items share the same relation.
- Do not over-generalize beyond the evidence.
- If clusters are provided, merge only within the same cluster.

Output schema:
{
  "consolidated_events": [
    {
      "consolidated_text": "string",
      "memory_kind": "stable_fact|preference|ongoing_plan|relationship|recurring_theme|status|other",
      "time_range": [number, number],
      "participants": ["string", ...],
      "entity_hints": ["string", ...],
      "source_event_ids": ["string", ...],
      "stability": "stable|evolving|conflicting",
      "invariants": ["string", ...],
      "evolution": ["string", ...],
      "conflicts": [
        {
          "claim_a": "...",
          "claim_b": "...",
          "source_event_ids_a": [...],
          "source_event_ids_b": [...]
        }
      ],
      "confidence": number
    }
  ],
  "unmerged_event_ids": ["string", ...]
}
\end{promptbox}

\subsection{Evaluation Prompts}
\label{app:evaluation_prompts}

\begin{promptbox}{LLM-Judge prompt}
Your task is to label an answer to a question as "CORRECT" or "WRONG".
You will be given:
(1) a question,
(2) a gold answer,
(3) a generated answer.

The question asks about something one user should know about another user based on prior conversations.
The gold answer is usually concise.
The generated answer may be longer, but should be counted as CORRECT if it refers to the same fact or topic as the gold answer.

For time-related questions, consider different surface forms of the same date or time period as CORRECT.
For example, "May 7th" and "7 May" should be considered equivalent if they refer to the same date.

Question: {question}
Gold answer: {gold_answer}
Generated answer: {generated_answer}

Return ONLY valid JSON:
{"label": "CORRECT"} or {"label": "WRONG"}.
\end{promptbox}

\begin{promptbox}{Answer simplification prompt}
You are an answer simplifier.
You will be given a question and a generated answer.
Extract only the minimal core fact that directly answers the question.

Rules:
- Do NOT add explanations or extra context.
- Preserve names, numbers, dates, and units when they are part of the answer.
- If the generated answer contains multiple facts, keep only the fact required by the question.
- Output MUST be valid JSON only.

JSON schema:
{"answer": <string>}

Question: {question}
Generated Answer: {answer}
\end{promptbox}

\subsection{Retrieval Planning Prompts}
\label{app:retrieval_planning_prompts}

\begin{promptbox}{Subquery planner prompt}
You are to decompose the user's query into atomic subqueries and their dependencies.
Your goal is to produce subqueries that are SMALLER, MORE CONSTRAINED, and SEMANTICALLY SINGLE-PURPOSE than the original query, never broader.

You must think silently and output ONLY ONE JSON object.
No chain-of-thought, no analysis, no <think>, no markdown, no code fences.

JSON schema:
- "subqueries": a list of objects each with:
    - "id": "q1","q2",...
    - "text": string (the atomic subquery)
    - "memory_scope": one of ["SHORT","LONG","MIXED"]
    - "coverage_mode": one of ["local","global"]
    - "type_hint": optional short string hint for unit type (e.g., "intention","procedural","preference","episodic","semantic"); may be null
    - "deps": list of subquery ids this subquery depends on (may be empty)
    - "hint_time": null or [start_timestamp, end_timestamp] in seconds if time-constrained
- "dependency_graph": object mapping subquery id -> list of dependency subquery ids (prerequisites).

CRITICAL PRINCIPLE (DO NOT VIOLATE):
- Each subquery must preserve ALL explicit constraints from the user query that are relevant to answering it.
- NEVER drop or weaken constraints such as:
  - time windows / dates / relative-time anchors
  - named entities / subjects / objects
  - locations
  - quantities / counts
  - qualifiers like "major", "first", "most recent", "in that meeting", "during that trip", etc.
- If a constraint exists in the user query, it MUST appear in the subquery text OR be represented as hint_time when appropriate.
- Subqueries must not broaden the search space. They should reduce it.

TIME WINDOW RULES:
- If the query includes a clear absolute time (e.g., a month/year/date), keep it in the subquery text.

DECOMPOSITION RULES:
- Prefer 1-5 subqueries. If the user query is already atomic and well-scoped, output exactly 1 subquery that restates it faithfully (do NOT split).
- HARD RULE: If you output EXACTLY 1 subquery, its "text" MUST be EXACTLY IDENTICAL to the full original user query (verbatim). Do NOT paraphrase, shorten, translate, or reformat.
- Decompose only when it reduces difficulty by isolating distinct required pieces (e.g., identify entity, then retrieve evidence, then synthesize).
- When decomposing, subqueries should become more specific via:
  - separating different targets (different people/items/events) into separate subqueries, while preserving the original constraints for each target
  - separating different time windows or temporal anchors into separate subqueries, especially when one date/time/event must be used as the reference point for another step
  - separating anchor resolution from calculation/comparison/synthesis: first retrieve the date/entity/event that acts as the anchor, then use it in dependent subqueries
- NEVER create a subquery that is a strict superset of another subquery's scope unless it is explicitly needed for global coverage questions.

ATOMICITY & SEMANTIC PURITY (VERY IMPORTANT):
- EACH subquery must represent exactly ONE information need / ONE intent / ONE semantic predicate.
- A subquery must be answerable by ONE short response (or one focused list) without needing to address a second, separate question.
- DO NOT combine multiple asks in one subquery. Avoid conjunction-style bundling such as:
    - "and", "or", "as well as", "also", "+"
- If the user query contains multiple intents, split them into multiple subqueries.
- If a question contains "what" + "why" + "how" (or similar), split:
    - e.g., (1) what happened? (2) why did it happen? (deps=[(1)]) (3) how was it decided? (deps=[(1)])
- If the query asks for comparison or trade-off (A vs B), prefer:
    - one subquery to retrieve evidence about A,
    - one subquery to retrieve evidence about B,
    - optional synthesis subquery "Compare A vs B based on evidence" with deps on both.
- If you find yourself writing a long subquery with multiple clauses, split it.

DEPENDENCIES:
- Use deps only for true prerequisites (e.g., resolve an ambiguous referent before retrieving evidence).
- Do NOT produce circular dependencies: ensure the dependency_graph is acyclic.

COVERAGE MODE (local vs global):
- Keep coverage_mode="local" for single-incident, single-moment questions that can be answered from a small number of best-matching snippets.
- Use coverage_mode="global" whenever answering requires combining evidence from MULTIPLE turns.
    This includes aggregation intents even if the user does NOT explicitly say "all"/"so far".
    MUST set coverage_mode="global" for any counting/aggregation intent, e.g.:
        - "count ...", "how many ...", "number of ...", "total ...", "times ...", "frequency ...", "rate ..."
    Also set coverage_mode="global" for set-collection intents, e.g.:
        - "list ...", "names of ..."

MEMORY SCOPE:
- SHORT: use for episodic or situational dialogue evidence tied to a specific moment, date, meeting, action, decision, or temporary context.
  Examples: "meeting is tomorrow at 3pm", "the user bought a tennis racket yesterday", "focus on page 4 of the uploaded file".

- LONG: use for stable, reusable, or profile-like memory that remains useful beyond the original dialogue moment.
  such as enduring preferences, personal background, identity, education, occupation, relationships, hobbies, long-term goals, persistent statuses, and general procedures.
  Examples: "the user prefers concise answers", "the user graduated with a degree in Business Administration", "the user's favorite running shoe brand is Nike".

- MIXED: use when answering the query requires both LONG-style stable memory and SHORT-style situational evidence.
  Example: "the user generally prefers concise writing, but this specific email to the CEO should be more formal".
OUTPUT MUST be valid JSON following the schema above. Do NOT include any explanation, analysis, or markdown.

User query:
{QUERY_TEXT}
\end{promptbox}

\begin{promptbox}{Missing-information query prompt}
Return EXACTLY ONE compact single-line JSON object.
No chain-of-thought, no analysis, no <think>, no markdown, and no code fences.
Required key: missing_info_query.

You are generating ONE actionable retrieval query to fill the missing slot(s) AFTER a reasoner has already produced a preliminary answer.

Input is a JSON object with keys:
query (string),
subquery (string), and
reason_output (object with keys conclusions, confidence, and missing_info).

Missing-info policy:
missing_info MUST be true whenever the available evidence cannot answer the subquery as asked with the required specificity.
This includes, but is not limited to:
(a) the question asks for a specific value, but the evidence only provides a vague reference;
(b) the answer requires resolving an ungrounded pronoun or placeholder;
(c) there are multiple plausible candidates.

Anti-echo rule:
The missing-information query MUST NOT be a paraphrase or rewrite of the subquery.
It must ask for a missing slot explicitly suggested by the evidence but not specified.
It must include at least one concrete anchor from the evidence, such as a named entity, object, or time phrase.
If the subquery asks for X but the evidence only mentions Y, ask about Y first as a bridge step.
The query should be minimal and ordered by dependency.

Example:
Subquery: Which airline did John fly to Tokyo?
Evidence: John said, ``I took a red-eye to Tokyo last night.''
Good: {"missing_info_query":"Which airline operated the red-eye flight John took?"}
Bad: {"missing_info_query":"Which specific airline did John fly to Tokyo?"}

Output JSON only.
\end{promptbox}
\end{document}